\documentclass[]{TEAI}
\usepackage{helvet}

\usepackage{amsmath} 
\usepackage{natbib}
\usepackage{graphicx}
\usepackage{subcaption} 

\usepackage[toc,page,header]{appendix}
\usepackage[utf8]{inputenc} 
\usepackage[T1]{fontenc}    
\usepackage{hyperref}       
\usepackage{url}            
\usepackage{booktabs}       
\usepackage{lmodern}        
\usepackage{amsfonts}       
\usepackage{nicefrac}       
\usepackage{microtype}      
\usepackage{wrapfig}

\usepackage{amssymb}  
\usepackage{fontawesome}  
\usepackage{url}  

\usepackage{titletoc}

\usepackage{tikz}  
\usepackage{comment}  
\usepackage{tabularx}  
\usepackage{booktabs}  

\usepackage{minitoc}

\usepackage{booktabs}
\usepackage{array}
\usepackage{etoolbox}

\definecolor{lightblue}{RGB}{200, 230, 255}  
\definecolor{headerblue}{RGB}{150, 200, 255} 

\usepackage{pgfplots}
\usepackage[utf8]{inputenc} 
\usepackage[T1]{fontenc}    
\usepackage{hyperref}       
\usepackage{url}            
\usepackage{booktabs}       
\usepackage{amsfonts}       
\usepackage{nicefrac}       
\usepackage{microtype}      
\usepackage{xcolor}         
\usepackage{graphicx}
\usepackage{float}
\usepackage{comment}
\usepackage{multirow} 
\usepackage{amsmath} 
\usepackage{makecell} 
\usepackage{siunitx}  
\usepackage{tikz}
\usepackage{pgf-pie} 
\usepackage{subcaption}
\usepackage{wrapfig}
\usepackage[export]{adjustbox}

\usepackage{ragged2e}      
\usepackage{tabularx}       
\usepackage{array}          
\usepackage{caption}        
\usepackage{enumitem}
\usepackage{pifont}
\usepackage[hang,flushmargin]{footmisc} 

\usepackage{tcolorbox}

\usepackage{tcolorbox}    
\tcbuselibrary{breakable}  
\tcbuselibrary{skins}      

\usepackage{tabularx}
\usepackage{listings}

\usepackage{algorithm}
\usepackage{algorithmic}
\usepackage{alltt}

\newcommand{\mycomment}[1]{\hfill\textit{// #1}}

\title{Understanding Curriculum Learning in Large Language Models via Cross-Difficulty Optimization Dynamics}

\author{
    Zhikai Ding\textsuperscript{1},
    Ziyi Ye\textsuperscript{1,$\dagger$}
}

\affiliation[1]{\mbox{Fudan University}} 

\abstract{
\begin{abstract}

Curriculum learning has been widely adopted in the post-training of large language models by organizing training data from easy to hard. However, its effectiveness varies substantially across reasoning tasks, suggesting that no single curriculum is universally optimal and raising a fundamental question: what determines when curriculum learning works? In this paper, we answer this question by analyzing the optimization dynamics induced by different curriculum schedules.
We show that the transfer relationship between different difficulty levels characterizes the optimization dynamics induced by curriculum learning, which in turn explains the effectiveness of different curriculum schedules, and formalize this relationship as Relative Transfer, a principled measure of cross-difficulty knowledge transfer. Based on this measurement, we derive Transfer-aware Dynamic Curriculum Sampling (TDCS), which dynamically adjusts the sampling distribution according to the estimated transfer relationship throughout training. Extensive experiments on multiple reasoning benchmarks demonstrate that TDCS consistently outperforms representative scheduling strategies across different tasks, model scales, and training paradigms. More importantly, our work provides a unified optimization-based explanation of curriculum learning through cross-difficulty transfer.

\end{abstract}
}

\correspondence{\email{dingzhikai158@gmail.com}}

\begin{document}
\maketitle
\renewcommand{\thefootnote}{}
\footnotetext{$^*$Equal Contribution.\\$^\dagger$Corresponding authors.}
\renewcommand{\thefootnote}{\arabic{footnote}}


\vspace{-1.5em}

\section{Introduction}

Large language models (LLMs) have achieved remarkable success across a wide range of reasoning tasks through supervised fine-tuning (SFT). As post-training datasets continue to grow in both scale and diversity, how to effectively organize training data has become increasingly important for improving optimization performance and model generalization. Consequently, training data scheduling has emerged as a fundamental component of LLM post-training, aiming to determine the order in which training examples are presented throughout optimization \citep{xu2026learningrightpaceadaptive, wang2026schedulingllmreinforcementlearning,sachdeva2024traindataefficientllms,hu-etal-2026-fine}.

Among various scheduling strategies, Curriculum Learning (CL) \citep{10.1145/1553374.1553380,pmlr-v80-weinshall18a,9392296,zhang-etal-2026-beyond,10.1145/3589335.3641257}, which organizes training examples from easy to hard, is arguably the most widely adopted paradigm. Originally proposed in traditional machine learning, curriculum learning has demonstrated consistent optimization and generalization benefits across numerous learning problems \citep{ICLR2025_9c77f2ce}. Motivated by its success, many recent LLM post-training methods have incorporated curriculum learning by ranking training samples according to predefined difficulty metrics and progressively exposing the model to increasingly challenging examples \citep{yang2026ttcstesttimecurriculumsynthesis,xu-etal-2020-curriculum,ranaldi-etal-2024-language}. Consequently, fixed easy-to-hard curricula are commonly adopted as the default scheduling strategy in existing LLM post-training pipelines.

Although curriculum learning has been widely adopted,
its effectiveness varies substantially across reasoning tasks. Existing studies have documented this phenomenon through large-scale empirical evaluations across different models, tasks, and difficulty metrics, they demonstrate that curriculum learning may outperform random sampling in some scenarios while becoming ineffective or even detrimental in others \citep{kim2024strategicdataorderingenhancing,jia-etal-2026-makes,wu2025progressivemasterycustomizedcurriculum,xu2026learningrightpaceadaptive}. Although these studies reveal the limitations of fixed curriculum schedules, they primarily provide empirical observations and analyses of difficulty metrics, leaving a fundamental question unanswered: \emph{what determines whether curriculum learning is effective?} Without understanding the underlying mechanism, it remains difficult to design more effective curriculum strategies beyond empirical trial and error.

In this paper, we revisit curriculum learning by analyzing the optimization dynamics induced by different curriculum schedules. Rather than treating curriculum learning as a predefined training heuristic, we seek to understand the optimization mechanism that determines when and why a curriculum schedule succeeds. To this end, we first conduct a systematic empirical study across multiple reasoning benchmarks. Our results reveal that no fixed scheduling strategy consistently performs best across different reasoning tasks, suggesting that the effectiveness of curriculum learning is fundamentally task-dependent rather than universally optimal.

To answer this question, we investigate the optimization dynamics induced by curriculum learning. Specifically, we analyze how optimization on one difficulty level influences the optimization of other difficulty levels throughout training: optimizing one difficulty level may either facilitate or interfere with the optimization of others, and the overall transfer relationship determines whether a fixed curriculum schedule is effective. Based on a first-order optimization analysis, we formalize this transfer relationship as \emph{Relative Transfer}, a principled measure of cross-difficulty knowledge transfer. This analysis provides a unified explanation of when curriculum learning succeeds or fails across different reasoning tasks, going beyond previous empirical observations.

Building upon this theoretical understanding, we derive \textbf{Transfer-aware Dynamic Curriculum Sampling (TDCS)}, an adaptive curriculum learning framework that dynamically adjusts the sampling distribution according to the estimated transfer relationship throughout training. Instead of following a predefined easy-to-hard schedule, TDCS allocates training samples based on the estimated transfer benefits across difficulty levels. Extensive experiments demonstrate that the resulting sampling strategy consistently outperforms existing fixed scheduling strategies across multiple reasoning benchmarks and model scales, while further generalizing to downstream self-improvement settings.

Our main contributions are summarized as follows:

\begin{itemize}
    \item We conduct a systematic empirical study of curriculum learning for LLM reasoning and demonstrate that no fixed scheduling strategy consistently performs best across different reasoning tasks.

    \item We provide a theoretical explanation of when curriculum learning succeeds or fails by analyzing cross-difficulty knowledge transfer, and formalize this mechanism through Relative Transfer.

    \item Based on the proposed transfer analysis, we derive Transfer-aware Dynamic Curriculum Sampling (TDCS), which consistently outperforms existing fixed scheduling strategies across multiple reasoning benchmarks, model scales, and self-improvement settings.
\end{itemize}

\section{Related Work}
\subsection{Curriculum Learning}

Curriculum learning organizes training samples from easy to hard according to predefined difficulty measures to facilitate optimization \citep{10.1145/1553374.1553380,soviany2022curriculumlearningsurvey,JMLR:v21:20-212}. Numerous variants have since been proposed, including self-paced learning and mentor-guided curriculum design \citep{Jiang_Meng_Zhao_Shan_Hauptmann_2015,8278851,wang-etal-2026-dsmentor}.

Recently, curriculum learning has been widely adopted in LLM post-training. Existing methods construct curricula based on reasoning complexity, model confidence, training loss, or estimated sample difficulty, demonstrating improved performance on reasoning and instruction-following tasks \citep{kim2024strategicdataorderingenhancing,jia-etal-2026-makes,rampp2024doesdefinitiondifficultymatter,WONG2026108438,tao2026dynamic}. However, they mainly focus on designing difficulty metrics or curriculum schedules.

In contrast, our work studies the optimization mechanism underlying curriculum learning. Rather than proposing another predefined schedule, we explain when and why different curriculum schedules become effective through cross-difficulty optimization analysis.

\subsection{Adaptive Data Scheduling}

Instead of relying on a predefined curriculum, adaptive data scheduling dynamically adjusts the sampling distribution according to the optimization state throughout training~\cite{xia2024lessselectinginfluentialdata,Hammoudeh_2024,pmlr-v151-silva22a}. Existing methods estimate sample importance using signals such as training loss, uncertainty, gradient information, or reinforcement learning objectives, and adaptively allocate training resources to improve optimization efficiency \citep{pmlr-v267-wang25bm,NEURIPS2025_cea04322,NEURIPS2025_a59ff5f7,ye2025learning}.

While these methods improve training efficiency through adaptive sample selection, their scheduling decisions are primarily driven by optimization signals that reflect the current training state. In contrast, our work focuses on the transfer relationship between different difficulty levels, providing a transfer-based criterion for curriculum scheduling rather than relying solely on optimization heuristics.
\section{Empirical Observation}

\begin{table*}[t]
\centering

\begin{minipage}[t]{0.42\textwidth}
\centering
\caption{Difficulty definition for different tasks.}
\label{tab:difficulty_definition}
\begin{tabular}{lll}
\toprule
\textbf{Task} & \textbf{Difficulty Definition}  \\
\midrule
Sudoku  & Number of blank cells \\
KodCode & GPT pass rate \\
iGSM    & Reasoning steps \\
\bottomrule
\end{tabular}
\end{minipage}
\hfill
\begin{minipage}[t]{0.56\textwidth}
\centering
\setlength{\tabcolsep}{6pt}
\caption{Performance comparison of scheduling strategies.}
\label{tab:scheduling_results}
\begin{tabular}{lccc}
\toprule
\textbf{Task} & \textbf{Curriculum} & \textbf{Mix} & \textbf{Random} \\
\midrule
Sudoku & \textbf{0.205} & \underline{0.190} & 0.164 \\
iGSM   & 0.350 & \textbf{0.390} & \underline{0.353} \\
Code   & 0.580 & \underline{0.582} & \textbf{0.607} \\
\bottomrule
\end{tabular}
\end{minipage}

\end{table*}

\begin{figure*}[t]
\centering

\subfloat[Sudoku]{
    \includegraphics[width=0.31\textwidth]{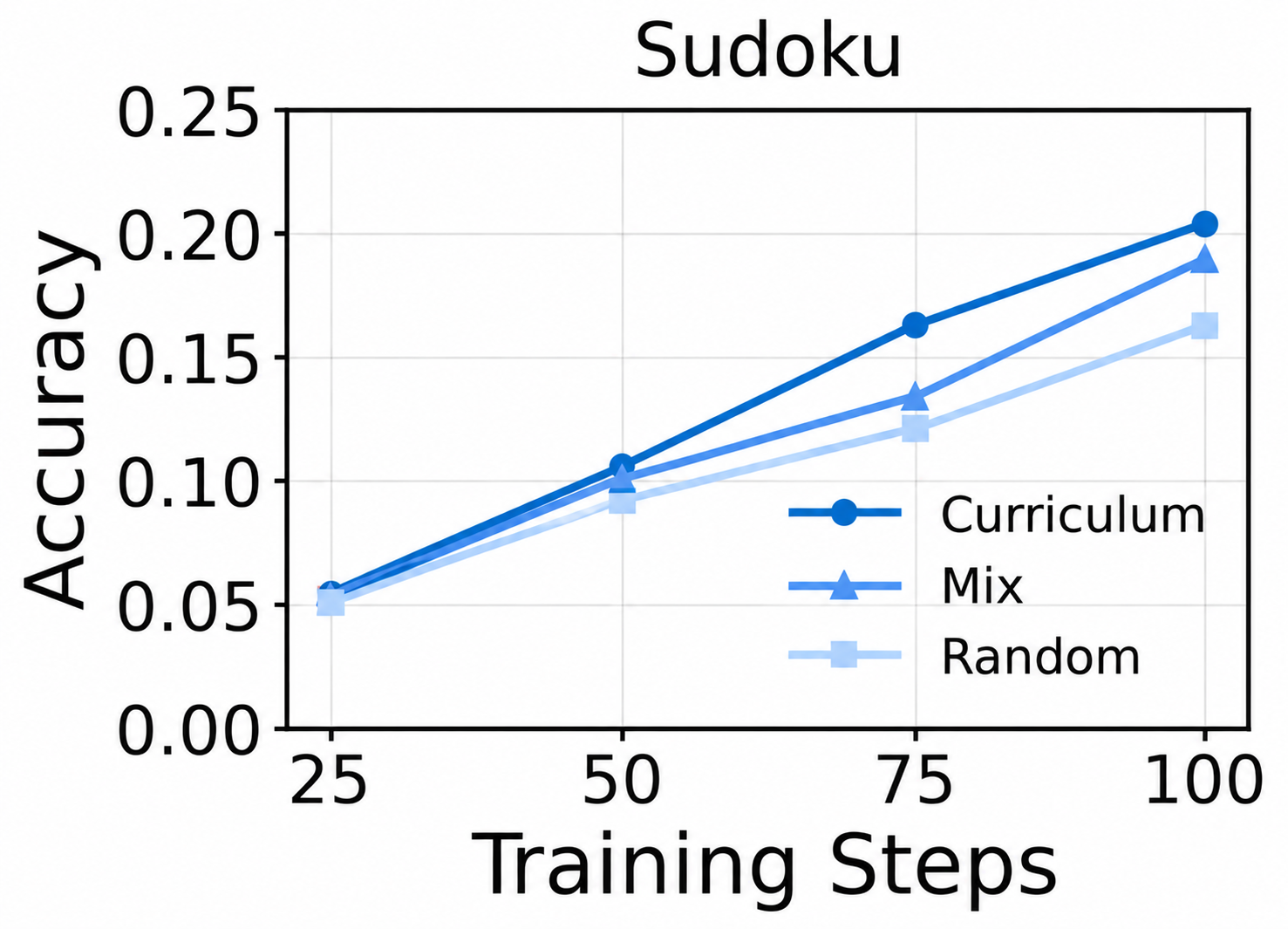}
}
\hfill
\subfloat[iGSM]{
    \includegraphics[width=0.31\textwidth]{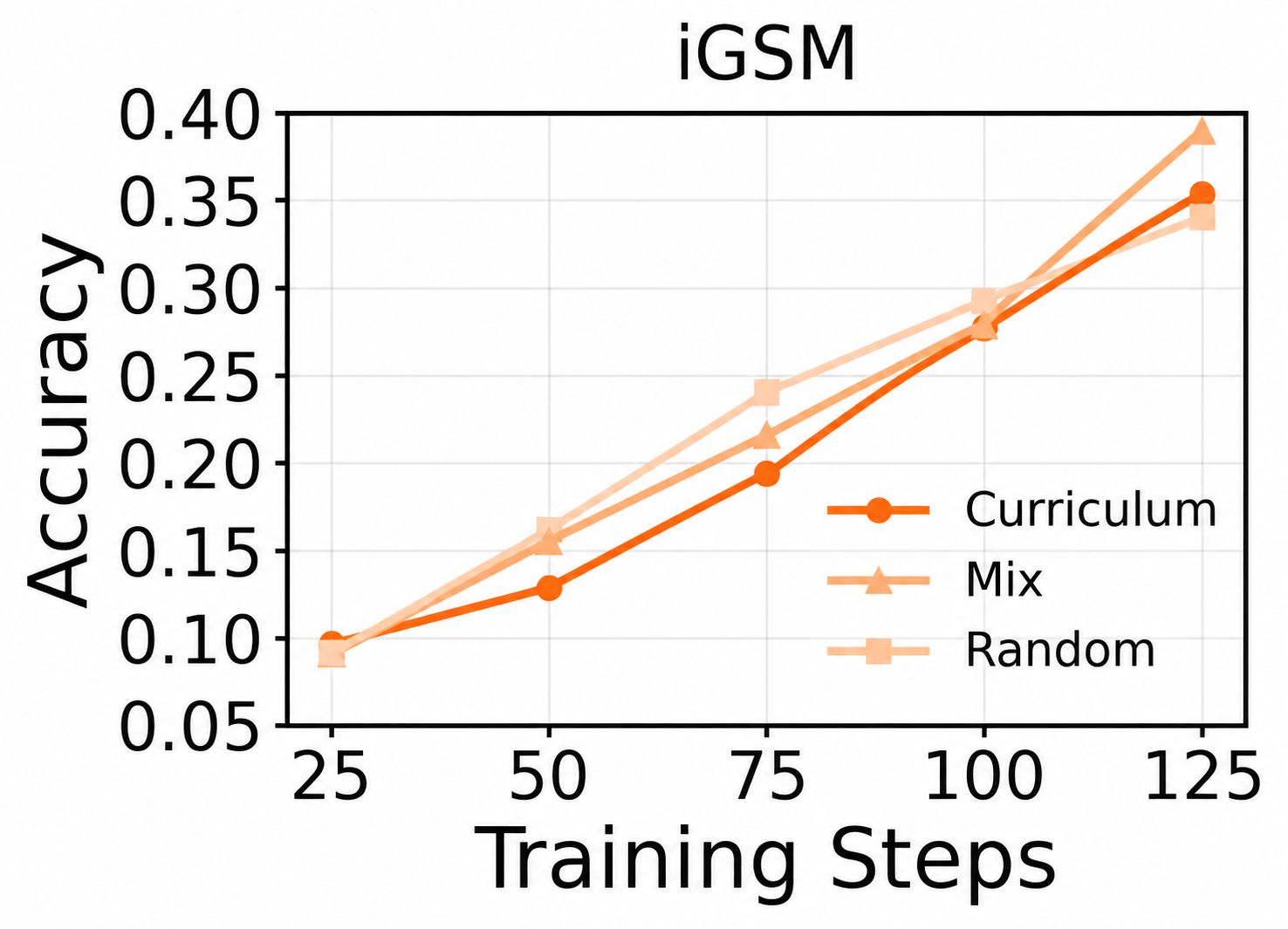}
}
\hfill
\subfloat[Code]{
    \includegraphics[width=0.31\textwidth]{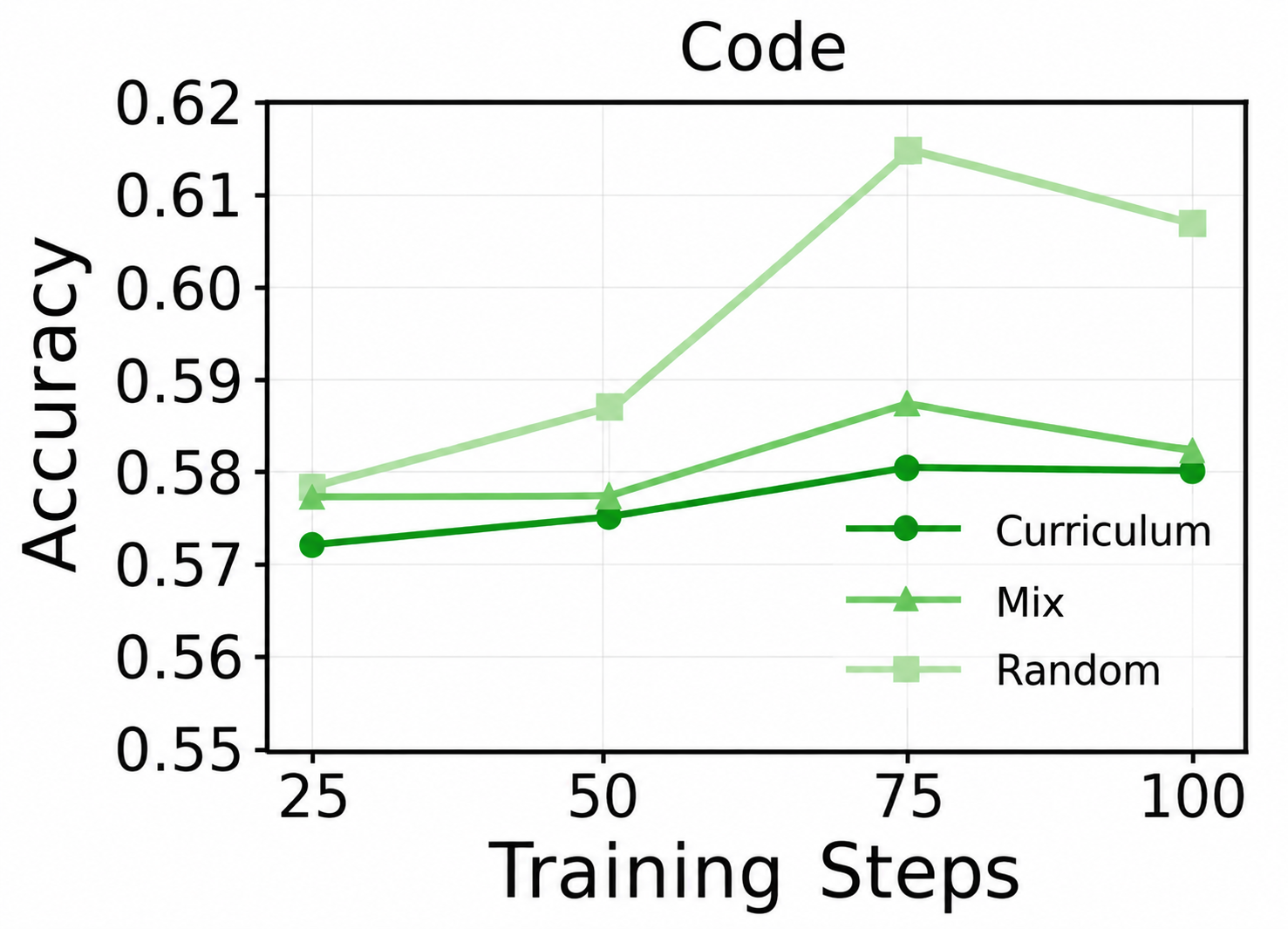}
}

\caption{Stepwise accuracy change of different schedules.}
\label{fig:three_tasks}
\end{figure*}

To investigate the effectiveness of curriculum learning in large language models, we conduct preliminary experiments on three reasoning benchmarks with naturally defined difficulty levels: Sudoku, KodCode \citep{xu-etal-2025-kodcode}, and iGSM \citep{ICLR2025_c239bac7}, covering logical reasoning, code generation, and mathematical reasoning, respectively.

Following the difficulty definitions in Table~\ref{tab:difficulty_definition}, the training data are partitioned into four difficulty levels for Sudoku and KodCode, and five difficulty levels for iGSM. We compare three representative training schedules: \textbf{Curriculum}, which trains the model from easy to hard; \textbf{Random}, which uniformly samples training examples from all difficulty levels; and \textbf{Mix}, which follows the curriculum schedule while replacing half of the samples at each stage with randomly sampled examples. All methods are trained under the same training budget.

Table~\ref{tab:scheduling_results} reports the final performance of the three scheduling strategies. Surprisingly, Curriculum does not consistently outperform the alternatives. Instead, Curriculum performs best on Sudoku, Mix achieves the highest accuracy on iGSM, and Random performs best on KodCode. These results indicate that curriculum learning is not universally beneficial for large language models, and no single scheduling strategy consistently achieves the best performance across different reasoning tasks.

The final performance, however, only reflects the optimization outcome. To obtain a more complete view, we further compare the training dynamics of different scheduling strategies in Figure~\ref{fig:three_tasks}. The three tasks exhibit markedly different optimization behaviors. While a single scheduling strategy consistently dominates throughout training on Sudoku and KodCode, the optimal strategy on iGSM changes as training progresses.

These observations suggest that the effectiveness of curriculum learning cannot be fully explained by the curriculum schedule itself. Instead, they raise an important question: what optimization mechanism determines whether a curriculum schedule succeeds? To answer this question, we next investigate the optimization process induced by curriculum learning, focusing on how optimization on one difficulty level influences the optimization of others.
\section{Optimization Analysis of Curriculum Learning}

\subsection{Modeling Knowledge Transfer}

The empirical observations in Section~3 show that no single scheduling strategy consistently performs best across different reasoning tasks. Understanding the underlying optimization mechanism is therefore essential for explaining the effectiveness of curriculum learning.

In this section, rather than analyzing the curriculum schedule itself, we investigate the optimization dynamics induced by curriculum learning. Specifically, we study how optimization on one difficulty level affects the optimization objective of another. We refer to this cross-difficulty interaction as knowledge transfer.

Assume that the current optimization step is performed on examples from difficulty level $j$. We are interested in quantifying how this update affects the loss of another difficulty level $i$. Such cross-difficulty interaction characterizes the knowledge transfer during curriculum learning.

Let the model parameter at training step $t$ be denoted by $w_t$. After performing one gradient descent step on data with difficulty level $j$, the model is updated as

\begin{equation}
w_{t+1}=w_t-\eta g_j,
\end{equation}

where $\eta$ is the learning rate and $g_j=\nabla L_j(w_t)$ is the gradient computed on difficulty level $j$.

We then analyze the loss of difficulty level $i$ after this update. Applying the first-order Taylor expansion gives

\begin{equation}
L_i(w_{t+1})
\approx
L_i(w_t)
+
\nabla L_i(w_t)^T
(w_{t+1}-w_t).
\end{equation}

Since

\begin{equation}
\nabla L_i(w_t)=g_i,
\end{equation}

substituting Eq.~(1) into Eq.~(2) yields

\begin{equation}
L_i(w_{t+1})
\approx
L_i(w_t)
-
\eta g_i^Tg_j.
\end{equation}

Therefore, the expected loss variation on difficulty level $i$ after optimizing difficulty level $j$ can be approximated by

\begin{equation}
\Delta L_i
=
L_i(w_{t+1})-L_i(w_t)
\approx
-\eta g_i^Tg_j.
\end{equation}

Eq.~(5) shows that the influence of optimizing one difficulty level on another is determined by the projection of one gradient onto the descent direction induced by the other. When $g_i^Tg_j>0$, optimizing difficulty level $j$ reduces the loss of difficulty level $i$, indicating positive knowledge transfer. In contrast, negative values imply that optimizing one difficulty level increases the loss of another, resulting in optimization conflicts.

However, the quantity $g_i^Tg_j$ alone does not provide a fair comparison of transfer across different target difficulty levels, since its magnitude is also influenced by the scale of the target gradient $g_i$. To compare transfer effects independently of gradient magnitude, we measure the loss reduction achieved by optimizing difficulty level $j$ \emph{relative} to that achieved by directly optimizing difficulty level $i$ itself.

When optimizing the target difficulty level $i$, Eq.~(5) gives an expected loss reduction proportional to $g_i^Tg_i$. In contrast, optimizing difficulty level $j$ reduces the same loss by an amount proportional to $g_i^Tg_j$. Their ratio therefore naturally measures the relative transfer effect from difficulty level $j$ to difficulty level $i$, which we define as the \textbf{Relative Transfer}

\begin{equation}
\mathrm{Re}(i,j)
=
\frac{g_i^Tg_j}
{g_i^Tg_i}.
\end{equation}

Intuitively, $\mathrm{Re}(i,j)$ measures the effectiveness of optimizing difficulty level $j$ relative to directly optimizing difficulty level $i$. A value of $\mathrm{Re}(i,j)=1$ indicates that optimizing difficulty level $j$ is expected to reduce the loss of difficulty level $i$ as much as directly optimizing $i$ itself. Values between $0$ and $1$ indicate partial positive transfer, values larger than $1$ indicate even stronger transfer than self-optimization, while negative values imply optimization conflicts.

\subsection{Transfer Analysis}

\begin{figure*}[t]
    \centering
    \includegraphics[width=0.65\textwidth]{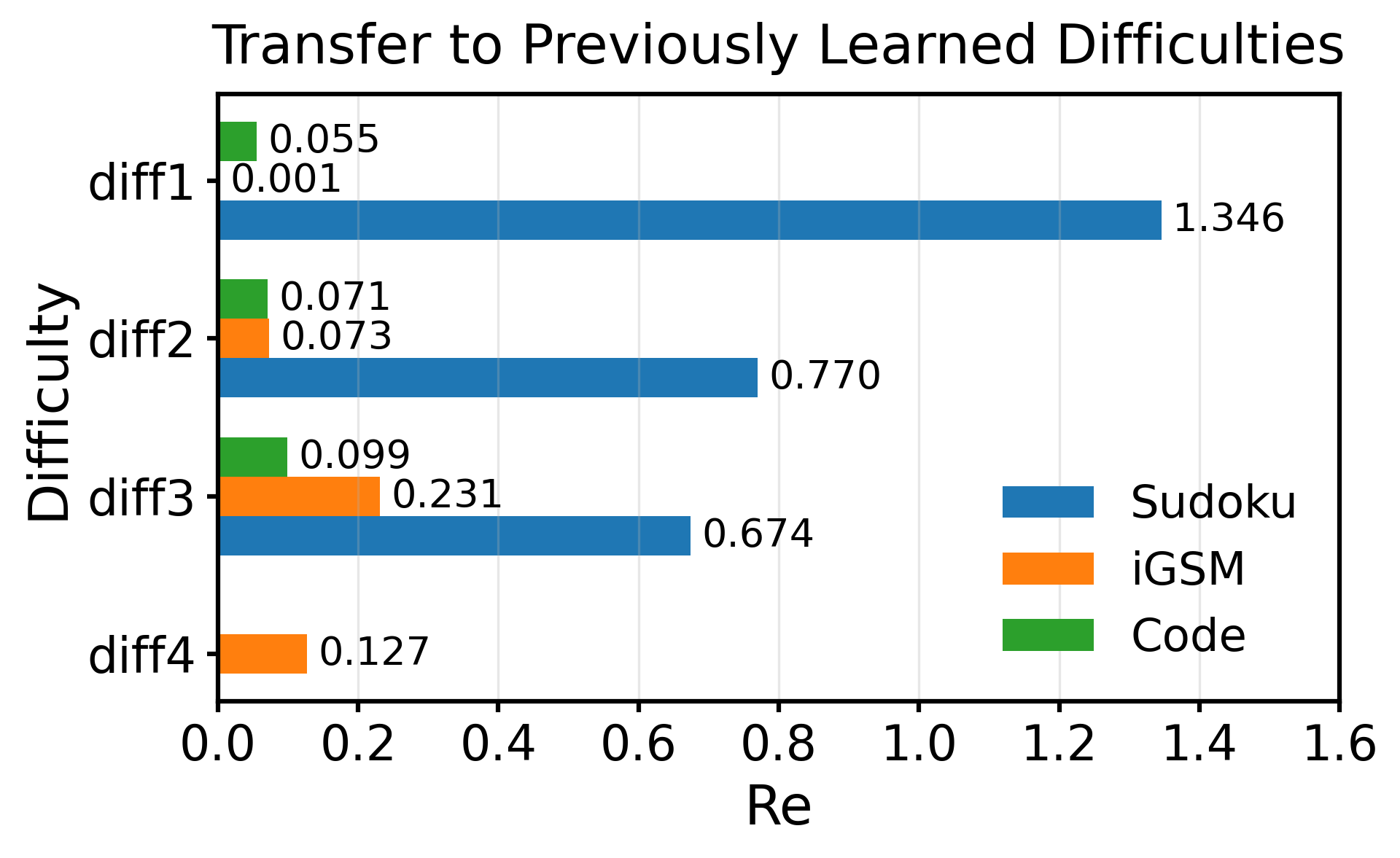}
    \caption{Transfer from the current difficulty to other difficulties. Sudoku exhibits consistently strong hard-to-easy transfer, while iGSM and Code show much weaker transfer.}
    \label{fig:transfer}
\end{figure*}

Eq.~(6) characterizes the knowledge transfer between arbitrary difficulty levels. In the following analysis, we focus on the final curriculum stage and examine the transfer from the current hardest difficulty level to all previously learned difficulty levels.

Figure~\ref{fig:transfer} visualizes the transfer from the current hardest difficulty level to all previously learned difficulty levels on each task. Sudoku exhibits consistently strong positive transfer, whereas iGSM and KodCode show much weaker transfer to earlier difficulty levels.

These transfer patterns provide an explanation for the empirical observations in Section~3. For Sudoku, strong transfer from the hardest difficulty continues to reduce the losses of previously learned difficulties. Consequently, optimizing only the hardest examples is sufficient to maintain performance across all difficulty levels, making a pure curriculum schedule effective. In contrast, the much weaker transfer observed on iGSM and KodCode suggests that optimizing only the hardest examples is insufficient to maintain performance on earlier difficulties, motivating the inclusion of additional replay from previous difficulty levels.

These findings suggest that the optimal sampling strategy should adapt to the observed cross-difficulty transfer relationship rather than follow a predefined curriculum schedule. This observation directly motivates the transfer-aware dynamic curriculum sampling strategy introduced in the next section.
\section{Transfer-aware Dynamic Curriculum Sampling}

Based on the transfer analysis in Section~4, the optimal curriculum should adapt to the observed cross-difficulty transfer relationship rather than follow a predefined easy-to-hard schedule. This observation naturally leads to three design principles. First, the amount of training allocated to the current difficulty should increase with its transfer capability. Second, replay should focus on difficulty levels that receive insufficient transfer. Third, when a harder difficulty provides exceptionally strong positive transfer, it should also be incorporated into training. Based on these principles, we derive Transfer-aware Dynamic Curriculum Sampling (TDCS). The overall procedure is summarized in Algorithm 1.

\paragraph{Current Difficulty Ratio Mapping.}

Let the current training stage correspond to difficulty level $k$, referred to as the \emph{current difficulty} throughout this section. The first step is to determine how much training should remain on the current difficulty. This directly follows from the transfer analysis in Section~4: if the current difficulty can effectively optimize previously learned difficulties, replay becomes less necessary. Intuitively, if the current difficulty can effectively transfer knowledge to previously learned difficulty levels, more training should remain on the current difficulty; otherwise, additional replay is required.

Since difficulty levels that can already be effectively optimized through knowledge transfer require little additional replay, we first identify the difficulty levels that receive insufficient transfer from the current difficulty. Specifically, we select the difficulty levels whose Relative Transfer is smaller than a threshold $\tau_e$,
\begin{equation}
\mathcal{S}=\{i\neq k \mid Re(i,k)<\tau_e\}.
\end{equation}

We then compute the average Relative Transfer
\begin{equation}
\bar{R}=\frac{1}{|\mathcal{S}|}\sum_{i\in\mathcal{S}}Re(i,k),
\end{equation}
which summarizes the overall transfer capability of the current difficulty to the difficulty levels that require additional replay.

Finally, the average transfer is mapped to the sampling ratio of the current difficulty through a sigmoid function,
\begin{equation}
\rho_k=\frac{1}{1+\exp[-\alpha(\bar{R}-\beta)]},
\end{equation}
where $\alpha$ and $\beta$ control the slope and midpoint of the mapping, respectively. The sigmoid function provides a smooth transition between replay-oriented and current-difficulty-oriented sampling. Consequently, stronger Relative Transfer leads to a larger sampling ratio for the current difficulty, while weaker transfer encourages more replay of previous difficulty levels.

\paragraph{Difficulty Allocation.}
Given the sampling ratio of the current difficulty, the remaining probability $(1-\rho_k)$ is allocated to the selected difficulty levels. To prioritize difficulty levels that receive weaker knowledge transfer, we adopt a reverse exponential weighting strategy,
\begin{equation}
p_i=(1-\rho_k)
\frac{\exp(-\lambda Re(i,k))}
{\sum_{j\in\mathcal{S}}\exp(-\lambda Re(j,k))},
\end{equation}
where difficulty levels with weaker transfer receive larger sampling probabilities, encouraging additional replay for difficulty levels that are less likely to benefit from the current optimization.

\paragraph{Harder Difficulty Adjustment.}
The previous steps determine the sampling distribution according to the estimated Relative Transfer. Nevertheless, when a harder difficulty provides exceptionally strong positive transfer, allocating additional training to that difficulty can further improve subsequent optimization. Therefore, if the Relative Transfer from the current difficulty to a harder difficulty exceeds a threshold $\tau_h$, i.e.,
\begin{equation}
Re(k,j)>\tau_h,
\end{equation}
a portion of the current sampling probability is reassigned to the harder difficulty in proportion to its Relative Transfer. Finally, all sampling probabilities are normalized to obtain the sampling distribution for the next training stage.












\begin{center}
\begin{minipage}{0.73\linewidth}

\hrule
\vspace{2pt}

\noindent
\textbf{Algorithm 1} \quad
\textbf{Transfer-aware Dynamic Curriculum Sampling}

\vspace{2pt}
\hrule
\vspace{3pt}

\begin{algorithmic}[1]

\REQUIRE Relative Transfer matrix $\mathbf{Re}$, current difficulty $k$
\ENSURE Sampling distribution $\mathbf{p}$

\STATE $\mathcal{S}\leftarrow\{i\neq k \mid Re(i,k)<\tau_e\}$
       \mycomment{Re Filter}

\STATE $\bar{R}\leftarrow
|\mathcal S|^{-1}\sum_{i\in\mathcal S}Re(i,k)$
       \mycomment{Mean Transfer}

\STATE $\rho\leftarrow\mathrm{Sigmoid}(\bar R)$
       \mycomment{Ratio mapping}

\STATE $p_k\leftarrow\rho$

\FORALL{$i\in\mathcal S$}
    \STATE $p_i\propto e^{-\beta Re(i,k)}$
           \mycomment{Difficulty allocation}
\ENDFOR

\FOR{$j>k$}
    \IF{$Re(k,j)>\tau_h$}
        \STATE $p_j\propto Re(k,j)$
               \mycomment{harder adjustment}
        \STATE $p_k\leftarrow p_k-p_j$
    \ENDIF
\ENDFOR

\STATE Normalize $\mathbf p$
\RETURN $\mathbf p$

\end{algorithmic}

\vspace{2pt}
\hrule

\end{minipage}
\end{center}

\begin{table*}[t]
\centering
\caption{Main results on three reasoning benchmarks.
The best result is shown in \textbf{bold}.}
\label{tab:main_results}
\small
\setlength{\tabcolsep}{10pt}
\begin{tabular}{llccc}
\toprule
\textbf{Category} & \textbf{Method} &
\textbf{Sudoku} &
\textbf{iGSM} &
\textbf{KodCode} \\
\midrule

\multirow{3}{*}{Fixed Schedule}
& Random      & 0.164 & 0.353 & 0.607 \\
& Curriculum  & 0.205 & 0.350 & 0.580 \\
& Mix         & 0.190 & 0.390 & 0.582 \\

\midrule

Adaptive Schedule
& \textbf{Ours}
& \textbf{0.231}
& \textbf{0.428}
& \textbf{0.619} \\

\bottomrule
\end{tabular}
\end{table*}

\begin{figure*}[t]
\centering

\subfloat[Sudoku]{
    \includegraphics[width=0.31\linewidth]{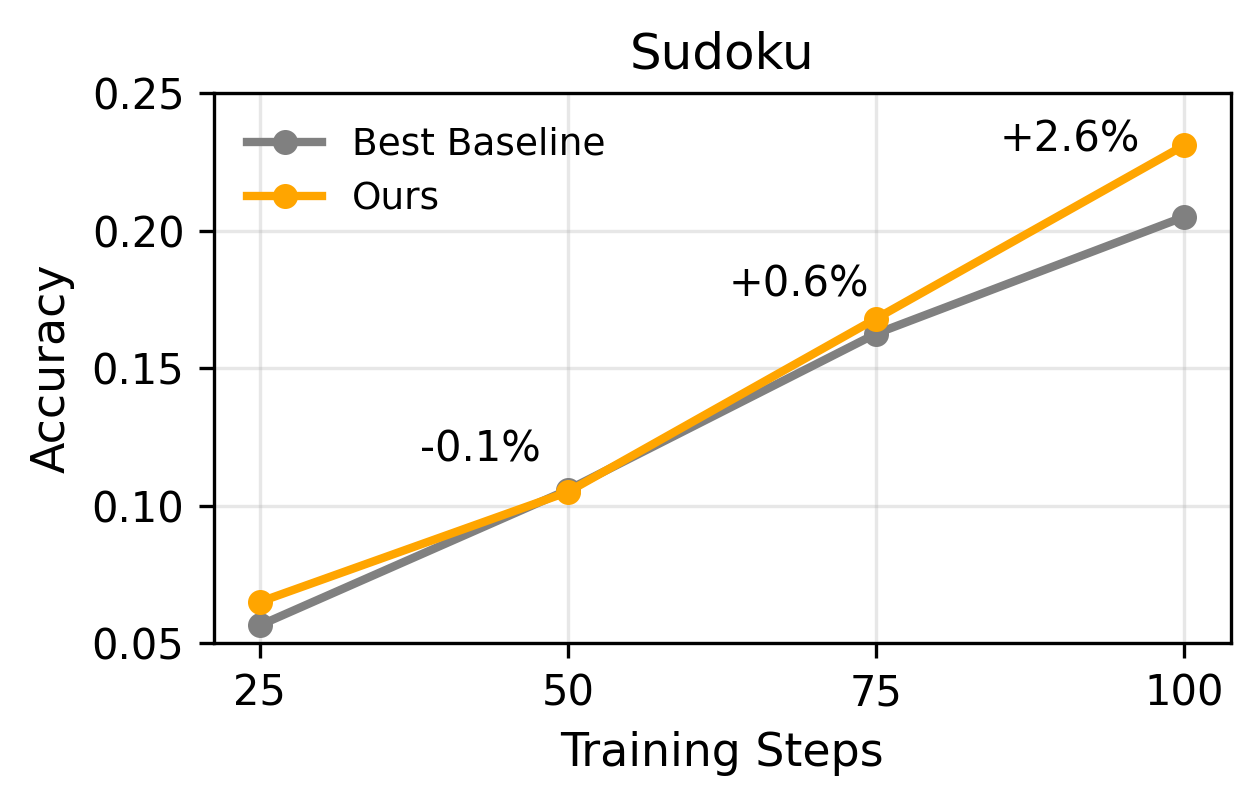}
    \label{fig:contrast_sudoku}
}
\hfill
\subfloat[iGSM]{
    \includegraphics[width=0.31\linewidth]{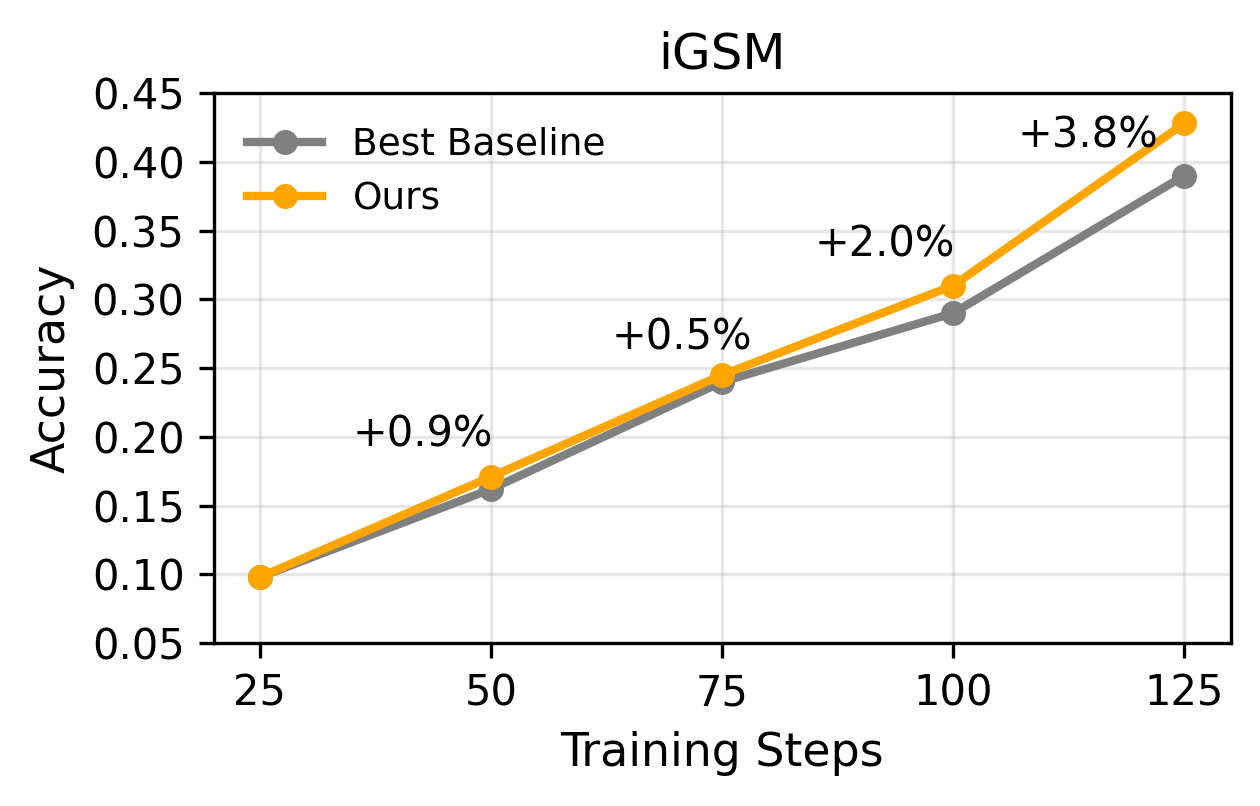}
    \label{fig:contrast_igsm}
}
\hfill
\subfloat[Code]{
    \includegraphics[width=0.31\linewidth]{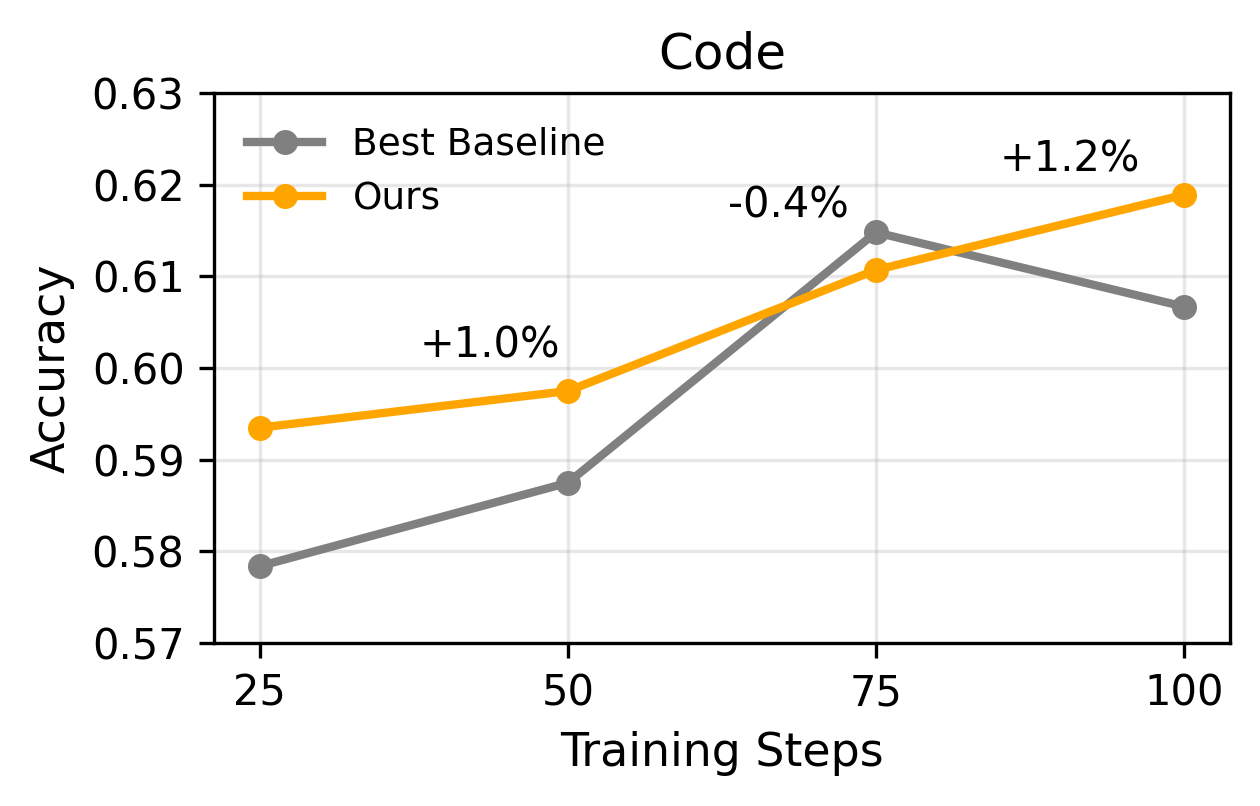}
    \label{fig:contrast_code}
}

\caption{
Training dynamics of the proposed method and the strongest fixed scheduling baseline during supervised fine-tuning.
}
\label{fig:contrast}
\end{figure*}
\section{Experiments}

\subsection{Experimental Settings}

\paragraph{Datasets.}
We evaluate our method on three reasoning benchmarks: \textbf{Sudoku}, \textbf{KodCode} \citep{xu-etal-2025-kodcode}, and \textbf{iGSM} \citep{ICLR2025_c239bac7}, covering logical reasoning, code generation, and mathematical reasoning, respectively. Following the difficulty definitions introduced in Section~3, Sudoku and KodCode are partitioned into four difficulty levels, while iGSM is partitioned into five difficulty levels. For self-improvement, we use GSM8K and KodCode dataset.

\paragraph{Models and Baselines.}
We conduct the main experiments on Qwen2.5-1.5B. To evaluate the generalization ability of the proposed method, we further consider Qwen2.5-3B, Qwen2.5-7B \citep{hui2024qwen25codertechnicalreport}, and Llama3.2-3B \citep{grattafiori2024llama3herdmodels}. We compare TDCS with three representative scheduling strategies introduced in Section~3: \emph{Curriculum}, \emph{Random}, and \emph{Mix}.

\paragraph{Implementation Details.}
All methods are trained under the same training data budget using LoRA fine-tuning. Unless otherwise specified, all methods are trained for 4 epochs with a learning rate of $1\times10^{-5}$ and a LoRA rank of 64. All remaining hyperparameters are kept identical across different methods to ensure a fair comparison.

\subsection{Main Results}

Table~\ref{tab:main_results} summarizes the final performance on the three reasoning benchmarks. Our method consistently achieves the best performance across all tasks, outperforming the strongest fixed scheduling strategy by 2.6\%, 3.8\%, and 1.2\% on Sudoku, iGSM, and KodCode, respectively. These results demonstrate that no single fixed curriculum is universally optimal, while dynamically adjusting the sampling distribution according to the estimated cross-difficulty transfer leads to consistently better optimization.

Figure~\ref{fig:contrast} further compares the training dynamics between TDCS and the strongest baseline. On Sudoku and iGSM, TDCS consistently maintains superior performance throughout training and converges to higher final accuracy. On KodCode, although the advantage is less pronounced during the intermediate stages, our method still achieves the best final performance. This observation suggests that transfer-aware curriculum adjustment not only improves the final optimization result but also provides a more stable training process across different reasoning tasks.

\begin{figure*}[t]
\centering

\subfloat[Sudoku]{
    \includegraphics[width=0.31\linewidth]{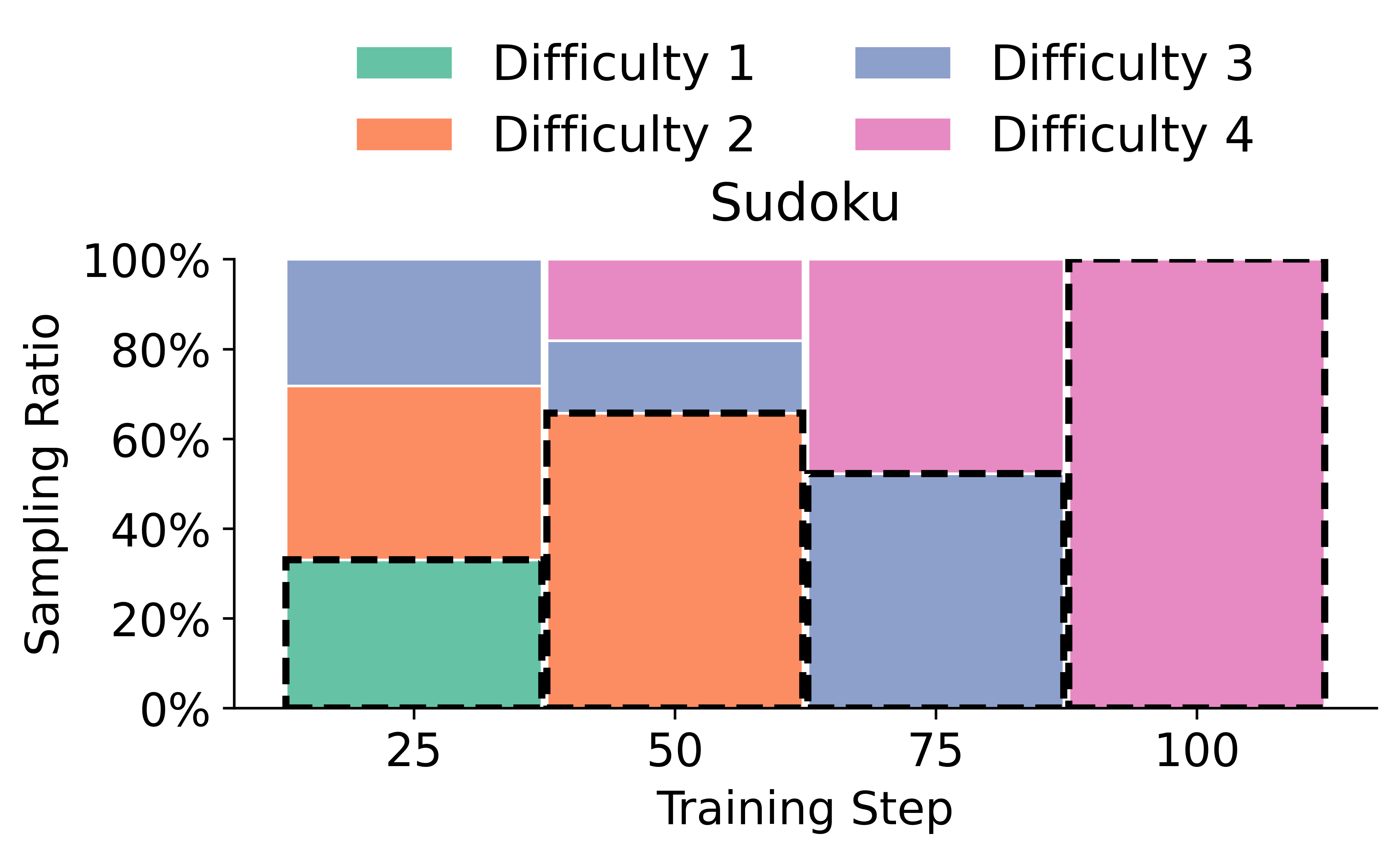}
    \label{fig:dist_sudoku}
}
\hfill
\subfloat[iGSM]{
    \includegraphics[width=0.31\linewidth]{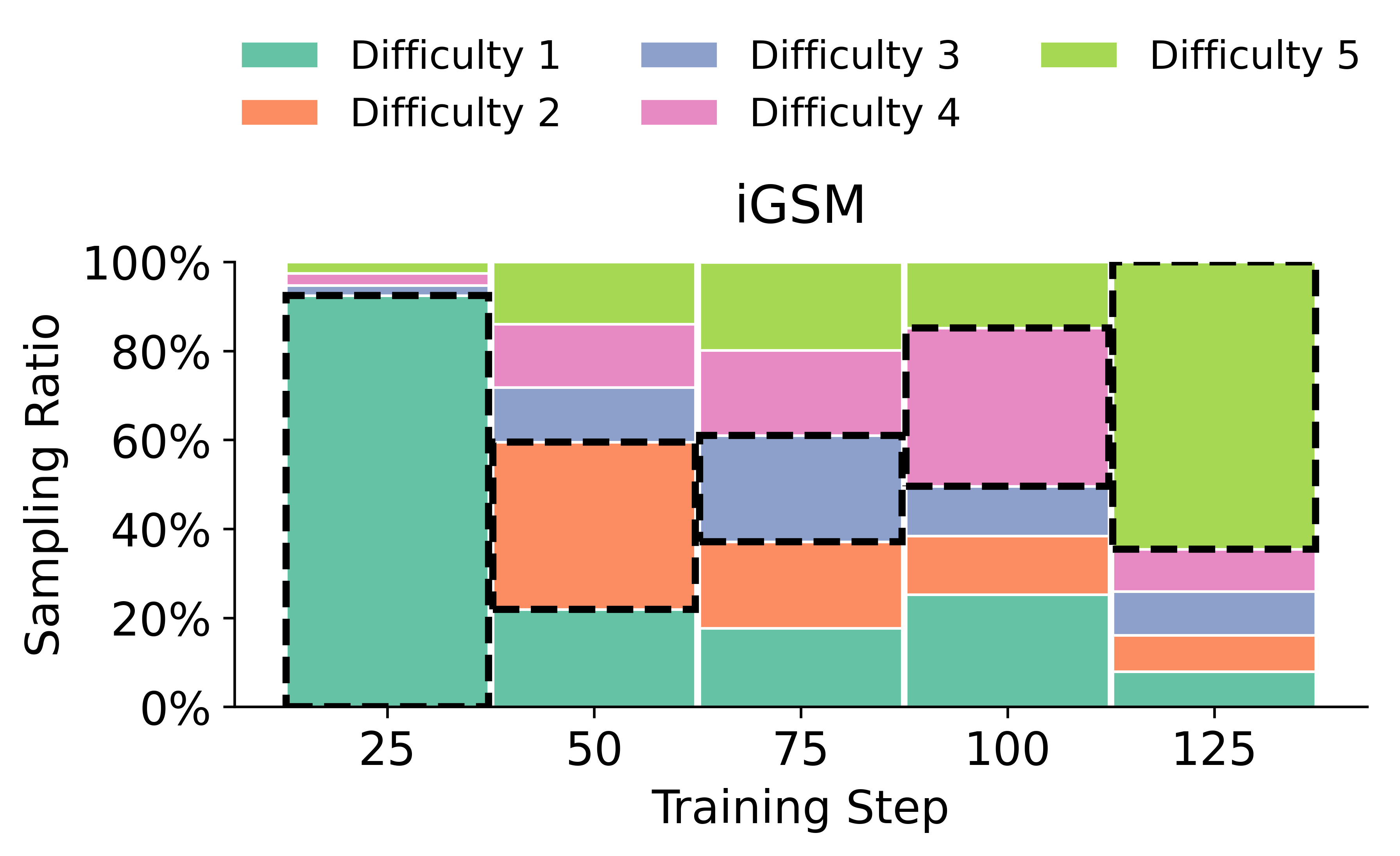}
    \label{fig:dist_igsm}
}
\hfill
\subfloat[Code]{
    \includegraphics[width=0.31\linewidth]{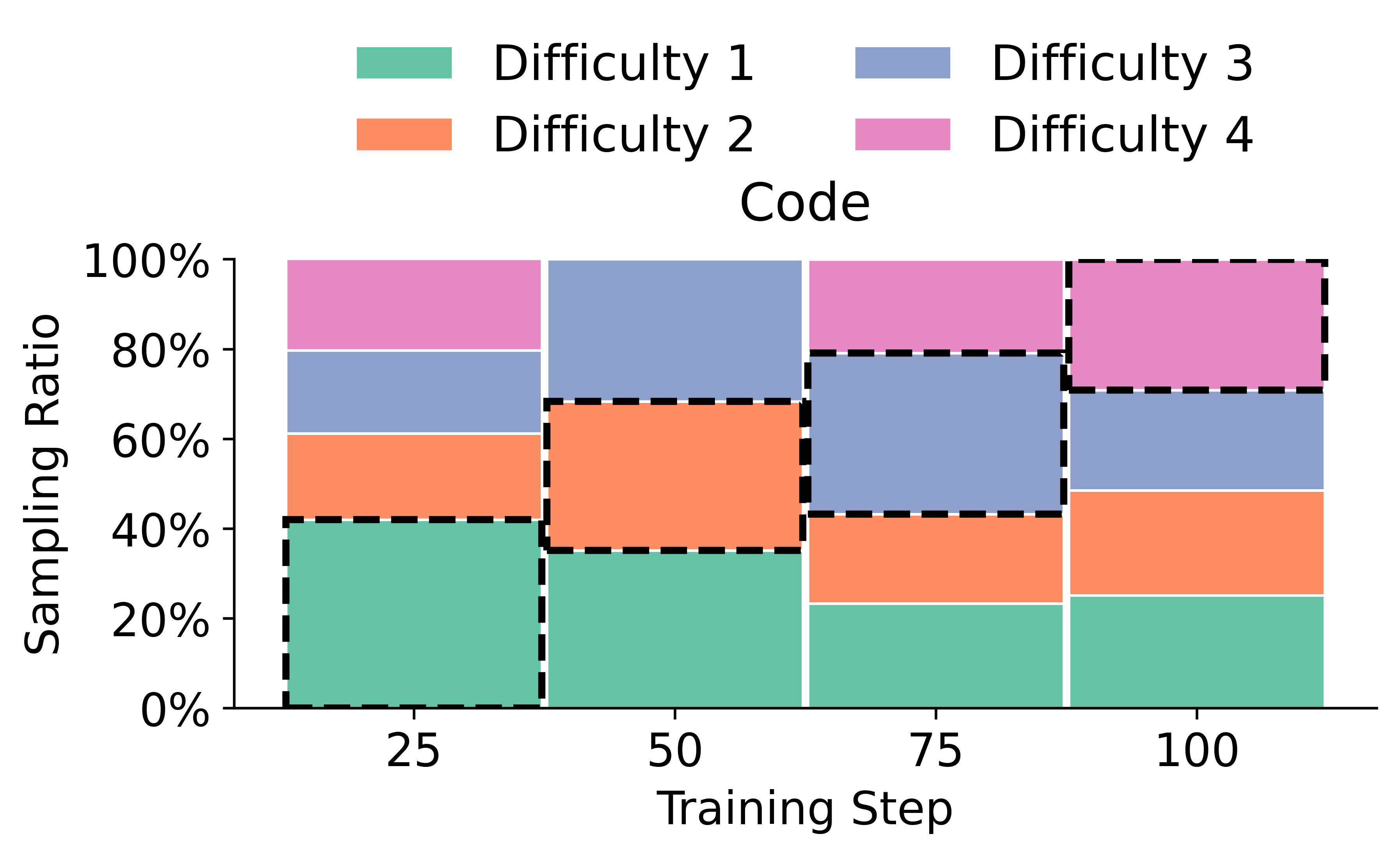}
    \label{fig:dist_code}
}

\caption{
Evolution of the sampling distribution produced by TDCS during supervised fine-tuning on the three reasoning benchmarks. Each stacked bar represents the sampling ratio assigned to different difficulty levels at a representative training stage, while dashed boxes indicate the current difficulty.
}
\label{fig:sampling_distribution}
\end{figure*}

To better understand how TDCS achieves these improvements, Figure~\ref{fig:sampling_distribution} visualizes the evolution of the sampling distribution during training. The sampling behavior differs substantially across reasoning tasks, reflecting their distinct transfer characteristics. Specifically, TDCS rapidly shifts the sampling distribution toward harder difficulties on Sudoku, where stronger cross-difficulty transfer reduces the need for replay. In contrast, iGSM maintains a moderate replay ratio throughout training, while KodCode preserves considerably larger replay ratios because of its weaker transfer capability. These observations are highly consistent with the transfer analysis in Section~4, demonstrating that TDCS dynamically adapts its curriculum according to the estimated Relative Transfer rather than following a predefined scheduling strategy.

\subsection{Generalization}

\paragraph{Generalization Across Model Scales.}

\begin{table*}[h]
\centering
\caption{Generalization across different model scales.
Numbers in parentheses denote the absolute accuracy improvement over the strongest baseline.}
\label{tab:generalization_models}

\small
\setlength{\tabcolsep}{6pt}

\begin{tabular}{lcccc}
\toprule
\multirow{2}{*}{\textbf{Model}}
& \multicolumn{2}{c}{\textbf{Sudoku}}
& \multicolumn{2}{c}{\textbf{iGSM}} \\
\cmidrule(lr){2-3}\cmidrule(lr){4-5}
& Baseline & Ours & Baseline & Ours \\
\midrule

Qwen2.5-3B
& 0.428
& \textbf{0.446 (+1.8\%)}
& 0.525
& \textbf{0.546 (+2.1\%)} \\

Qwen2.5-7B
& 0.438
& \textbf{0.497 (+5.9\%)}
& 0.766
& \textbf{0.819 (+5.3\%)} \\

Llama3.2-3B
& 0.405
& \textbf{0.433 (+2.8\%)}
& 0.673
& \textbf{0.875 (+20.8\%)} \\

\bottomrule
\end{tabular}
\end{table*}

To evaluate whether the proposed transfer-aware sampling strategy generalizes across different model scales and architectures, we further conduct experiments on Qwen2.5-3B-Instruct, Qwen2.5-7B-Instruct, and Llama3.2-3B-Instruct. As shown in Table~\ref{tab:generalization_models}, our method consistently outperforms the strongest fixed scheduling strategy on all evaluated models. Notably, the improvements are observed across both the Qwen and Llama model families, indicating that the proposed sampling strategy does not rely on a specific pretrained model or parameter scale. These results suggest that cross-difficulty knowledge transfer is a general optimization property, making the proposed transfer-aware curriculum broadly applicable to different LLMs.

\paragraph{Generalization Across Training Paradigms.}

\begin{table}[!h]
\centering
\small
\caption{Self-improvement results under TDCS.}
\label{tab:self_improvement}
\setlength{\tabcolsep}{4pt}
\begin{tabular}{lcccc}
\toprule
Task & Curriculum & Mix & Random & Ours \\
\midrule
GSM8K & \underline{0.612} & 0.601 & 0.601 & \textbf{0.622} \\
Code  & \underline{0.347} & 0.338 & 0.335 & \textbf{0.352} \\
\bottomrule
\end{tabular}
\end{table}

We further evaluate our method in a self-improvement setting, where the model is fine-tuned using its own generated responses instead of manually curated training data. In this setting, model answer accuracy is used as the difficulty metric to construct the curriculum. As shown in Table~\ref{tab:self_improvement}, our method consistently outperforms all fixed scheduling strategies on both GSM8K and KodCode. These results demonstrate that the proposed transfer-aware sampling strategy is compatible with different training paradigms and remains effective even when difficulty is estimated from the model's own predictions.




\subsection{Ablation Study}

\begin{table}[!h]
\centering
\small
\caption{Ablation study of TDCS components.}
\label{tab:ablation}
\setlength{\tabcolsep}{6pt}
\begin{tabular}{ccccc}
\toprule
HA & EDA & Sudoku & iGSM & Code \\
\midrule
$\times$ & $\checkmark$ & 0.213 & 0.416 & 0.612 \\
$\checkmark$ & $\times$ & 0.220 & 0.390 & 0.589 \\
$\checkmark$ & $\checkmark$ & \textbf{0.231} & \textbf{0.428} & \textbf{0.619} \\
\bottomrule
\end{tabular}
\end{table}

To evaluate the contribution of each component, we conduct ablation experiments on Qwen2.5-1.5B. In Table~\ref{tab:ablation}, \textbf{HA} denotes the \emph{Harder Adjustment} module, and \textbf{EDA} denotes the \emph{Exponential Difficulty Allocation} module, which replaces uniform probability allocation with transfer-aware exponential weighting over the selected difficulty levels.

As shown in Table~\ref{tab:ablation}, removing either component consistently degrades performance on all three benchmarks, confirming that both components contribute to the effectiveness of TDCS.

Removing \textbf{HA} results in the largest performance drop on Sudoku, where stronger positive transfer from harder to easier difficulty levels makes harder-difficulty adjustment particularly beneficial. In contrast, removing \textbf{EDA} consistently reduces performance across all tasks, indicating that allocating replay according to the estimated transfer relationship is more effective than uniform allocation. 

These results are consistent with the transfer analysis in Section~4 and demonstrate that both components are necessary for effectively exploiting cross-difficulty knowledge transfer.
\section{Conclusion}
In this paper, we revisited curriculum learning by analyzing the optimization dynamics induced by different curriculum schedules. We introduced \emph{Relative Transfer}, a principled measure that explains when curriculum learning succeeds by characterizing cross-difficulty knowledge transfer. Based on this analysis, we derived Transfer-aware Dynamic Curriculum Sampling (TDCS), which dynamically adjusts the sampling distribution according to the estimated transfer relationship and consistently outperforms representative curriculum scheduling strategies. We hope that the proposed transfer-based perspective provides a useful foundation for understanding curriculum learning. In future work, we plan to extend the proposed transfer analysis to broader optimization settings and explore more general transfer-aware data scheduling methods.
\section*{Appendix}
\section{Dataset Details}
\label{sec:appendix_dataset}

This section provides additional details of the datasets used in our experiments, including dataset statistics, difficulty definitions, and representative examples.

\subsection{Dataset Statistics}

Table~\ref{tab:dataset_statistics} summarizes the datasets used in this work. We report the number of training and evaluation samples, the number of difficulty levels, and the corresponding difficulty criterion for each benchmark.

\begin{table}[h]
\centering
\small
\caption{Statistics of the datasets used in this work.}
\label{tab:dataset_statistics}
\setlength{\tabcolsep}{4pt}
\begin{tabular}{lcccc}
\toprule
Dataset & Train & Test & \#Levels & Difficulty \\
\midrule
Sudoku & 800*4epochs & 1000 & 4 & Blank cells \\
iGSM & 1000*4epochs & 200 & 5 & Reasoning steps \\
KodCode & 800*4epochs & 200 & 4 & GPT pass rate \\
\bottomrule
\end{tabular}
\end{table}

\subsection{Dataset Examples}

This section shows representative examples from the three benchmarks.
For each example, we present the input, the expected output,
and its corresponding difficulty level.


\subsection{Sudoku Example}

\noindent
\fbox{%
\parbox{0.98\linewidth}{%

\textbf{Input}

\small

Fill the Sudoku. Replace each "?" with a digit from 1 to 9.

Rules:

Each row, each column, and each 3x3 box must contain the digits 1-9 exactly once.

Output ONLY the final completed 9x9 grid in the same format.

Grid:

[["?", 2, 7, 9, 3, 1, 4, 8, 5], [9, 3, 4, 5, 8, 6, 7, 1, 2], [1, 8, 5, 2, 7, "?", 6, "?", 3], [7, "?", "?", 4, 6, "?", "?", 3, 8], ["?", 4, 6, "?", 9, 8, 5, 2, "?"], [8, "?", 9, 3, 2, 7, 1, 6, 4], [5, 9, 8, 7, "?", 3, 2, 4, 6], [4, 6, 1, 8, "?", "?", 3, "?", 9], [2, "?", 3, 6, 4, 9, 8, 5, 1]]

\vspace{0.5em}

\textbf{Output}

[[6, 2, 7, 9, 3, 1, 4, 8, 5], [9, 3, 4, 5, 8, 6, 7, 1, 2], [1, 8, 5, 2, 7, 4, 6, 9, 3], [7, 1, 2, 4, 6, 5, 9, 3, 8], [3, 4, 6, 1, 9, 8, 5, 2, 7], [8, 5, 9, 3, 2, 7, 1, 6, 4], [5, 9, 8, 7, 1, 3, 2, 4, 6], [4, 6, 1, 8, 5, 2, 3, 7, 9], [2, 7, 3, 6, 4, 9, 8, 5, 1]]

\vspace{0.5em}

\textbf{Difficulty Level}: 1

}}


\subsection{iGSM Example}

\noindent
\fbox{%
\parbox{0.98\linewidth}{%

\textbf{Input}

\small

The number of each NetVersity's Microbiology Lab equals 4 times as much as each NetLearn Institute's Zoology Lab. The number of each NetVersity's Zoology Lab equals 19 more than each NetLearn Institute's Microbiology Lab. The number of each NetLearn Institute's Microbiology Lab equals 4. The number of each NetLearn Institute's Zoology Lab equals each NetVersity's Zoology Lab. How many Microbiology Lab does NetVersity have?

\vspace{0.5em}

\textbf{Output}

Define NetLearn Institute's Microbiology Lab as G; so G = 4. Define NetVersity's Zoology Lab as S; so S = 19 + G = 19 + 4 = 0. Define NetVersity's Microbiology Lab as P; so P = 4 * S = 4 * 0 = 0. Answer: 0

\vspace{0.5em}

\textbf{Difficulty Level}: 4

}}


\subsection{KodCode Example}

\noindent
\fbox{%
\parbox{0.98\linewidth}{%

\textbf{Input}

\small

def permute(nums):
"""
Generates all possible permutations of a given list using backtracking.

:param nums: List of elements to permute

:return: A list of all permutations

>>> permute([]) == [[]]

>>> permute([1]) == [[1]]

>>> permute([1, 2]) == [[1, 2], [2, 1]]

>>> result = permute([1, 2, 3])

>>> expected = [[1, 2, 3], [1, 3, 2], [2, 1, 3], [2, 3, 1], [3, 1, 2], [3, 2, 1]]

>>> sorted(result) == sorted(expected)

>>> result = permute([1, 1])

>>> expected = [[1, 1], [1, 1]]

>>> sorted(result) == sorted(expected)

"""

\vspace{0.5em}

\textbf{Output}

\small
\ttfamily

\begin{alltt}
def permute(nums):

    \hspace*{2em}"""
    
    \hspace*{2em}Generates all possible permutations of a \hspace*{2em} given list using
    backtracking.

    \hspace*{2em}:param nums: List of elements to permute
    
    \hspace*{2em}:return: A list of all permutations
    
    \hspace*{2em}"""
    
    \hspace*{2em}results = []

    \hspace*{2em}def backtrack(path):
    
        \hspace*{4em}if not nums:
        
            \hspace*{6em}results.append(path)
            
            \hspace*{6em}return

        \hspace*{4em}for i in range(len(nums)):
        
            \hspace*{6em}backtrack(path + [nums[i]])
            
            \hspace*{6em}nums.pop(i)

    \hspace*{2em}backtrack([])
    
    \hspace*{2em}return results
\end{alltt}

\vspace{0.5em}

\textbf{Difficulty Level}: 1
}}

\section{Training Details}
\label{sec:appendix_training}

This section summarizes the implementation details used in all experiments. We fine-tune the 7B model on 8 NVIDIA RTX 5090 GPUs, and all models smaller than 7B on 8 NVIDIA RTX 4090 GPUs.

\begin{table}[t]
\centering
\small
\caption{Training hyperparameters used in all experiments.}
\label{tab:training_details}
\begin{tabular}{lc}
\toprule
Hyperparameter & Value \\
\midrule

Fine-tuning method & LoRA \\

LoRA rank & 64 \\

LoRA alpha & 128 \\

LoRA dropout & 0.05 \\

Target modules & all \\

Optimizer & AdamW \\

Learning rate & 5e-5 \\

Scheduler & Constant \\

Epochs & 4 \\

Batch size / GPU & 4/8 \\

Gradient accumulation & 1 \\

Gradient checkpointing & Yes \\

Maximum sequence length & 512 \\

\bottomrule
\end{tabular}
\end{table}

During the training procedure of dynamic sampling curriculum learning, we sample and reweigh every difficulty per 25 steps, using 8 examples each difficulty level to calculate the Re matrix, which ensures evaluation accuracy while have a very low extra computation cost.

\subsection{Hyperparameter Sensitivity Analysis}

We conduct a sensitivity analysis of the two hyperparameters in our transfer-aware dynamic curriculum sampling strategy, namely $\tau_e$ and $\tau_h$. The threshold $\tau_e$ determines which easier difficulty levels are considered for replay, while $\tau_h$ controls when a harder difficulty level receives additional sampling probability based on its relative transferability. The results are reported in Table~\ref{tab:hyperparameter_sensitivity}.

The results reveal that the two hyperparameters affect different aspects of the sampling strategy. As shown in the left table, varying $\tau_e$ mainly affects the iGSM and Sudoku benchmarks. In particular, a lower $\tau_e$ changes the set of easier problems selected for replay, which has a more noticeable effect on tasks where replaying easier examples is beneficial. This effect is especially apparent on Sudoku, while the performance on Code remains relatively stable across the tested values.

In contrast, $\tau_h$ has a stronger influence on Sudoku, as shown in the right table. This is consistent with the characteristics of Sudoku, where the transfer relationship between easier and harder difficulty levels can be more pronounced. Increasing or decreasing $\tau_h$ changes when additional probability is allocated to harder difficulty levels, thereby directly affecting the balance between easier and harder examples.

Importantly, the middle configuration in each table corresponds to the hyperparameter values used by our final method, namely $\tau_e=0.5$ and $\tau_h=0.85$. These settings achieve consistently strong performance across all three benchmarks, suggesting that they provide a robust balance between replaying easier problems and allocating additional probability to harder problems. We therefore use $\tau_e=0.5$ and $\tau_h=0.85$ throughout the main experiments.

\begin{table}[t]
\centering
\caption{Sensitivity analysis of $\tau_e$ and $\tau_h$. The middle row in each table corresponds to the configuration used in our final method.}
\label{tab:hyperparameter_sensitivity}
\begin{minipage}{\linewidth}
\centering
\begin{minipage}[t]{0.48\linewidth}
\centering
\caption*{(a) Varying $\tau_e$}
\begin{tabular}{c|ccc}
\toprule
$\tau_e$ & iGSM & Code & Sudoku \\
\midrule
0.7 & 0.412 & 0.603 & 0.212 \\
0.5 & \textbf{0.428} & \textbf{0.619} & \textbf{0.231} \\
0.3 & 0.431 & 0.610 & 0.231 \\
\bottomrule
\end{tabular}
\end{minipage}
\hfill
\begin{minipage}[t]{0.48\linewidth}
\centering
\caption*{(b) Varying $\tau_h$}
\begin{tabular}{c|ccc}
\toprule
$\tau_h$ & iGSM & Code & Sudoku \\
\midrule
1.05 & 0.428 & 0.619 & 0.228 \\
0.85 & \textbf{0.428} & \textbf{0.619} & \textbf{0.231} \\
0.55 & 0.405 & 0.598 & \textbf{0.235} \\
\bottomrule
\end{tabular}
\end{minipage}
\end{minipage}
\end{table}

\clearpage

\bibliographystyle{plainnat}
\bibliography{main}

@inproceedings{ye2025learning,
  title={Learning llm-as-a-judge for preference alignment},
  author={Ye, Ziyi and Li, Xiangsheng and Li, Qiuchi and Ai, Qingyao and Zhou, Yujia and Shen, Wei and Yan, Dong and Liu, Yiqun},
  booktitle={International Conference on Learning Representations},
  volume={2025},
  pages={3537--3564},
  year={2025}
}

@misc{xu2026learningrightpaceadaptive,
      title={Learning at the Right Pace: Adaptive Data Scheduling Improves LLM Reinforcement Learning}, 
      author={Zicheng Xu and Ruixuan Zhang and Yu-Neng Chuang and Xiuyi Lou and Hoang Anh Duy Le and Oren Gal and Alexander S. Szalay and Zhaozhuo Xu and Guanchu Wang and Vladimir Braverman},
      year={2026},
      eprint={2606.22305},
      archivePrefix={arXiv},
      primaryClass={cs.CL},
      url={https://arxiv.org/abs/2606.22305}, 
}

@misc{wang2026schedulingllmreinforcementlearning,
      title={Scheduling Your LLM Reinforcement Learning with Reasoning Trees}, 
      author={Hong Wang and Zhezheng Hao and Jian Luo and Chenxing Wei and Yao Shu and Lei Liu and Qiang Lin and Hande Dong and Jiawei Chen},
      year={2026},
      eprint={2510.24832},
      archivePrefix={arXiv},
      primaryClass={cs.AI},
      url={https://arxiv.org/abs/2510.24832}, 
}

@misc{sachdeva2024traindataefficientllms,
      title={How to Train Data-Efficient LLMs}, 
      author={Noveen Sachdeva and Benjamin Coleman and Wang-Cheng Kang and Jianmo Ni and Lichan Hong and Ed H. Chi and James Caverlee and Julian McAuley and Derek Zhiyuan Cheng},
      year={2024},
      eprint={2402.09668},
      archivePrefix={arXiv},
      primaryClass={cs.LG},
      url={https://arxiv.org/abs/2402.09668}, 
}

@inproceedings{10.1145/1553374.1553380,
author = {Bengio, Yoshua and Louradour, J\'{e}r\^{o}me and Collobert, Ronan and Weston, Jason},
title = {Curriculum learning},
year = {2009},
isbn = {9781605585161},
publisher = {Association for Computing Machinery},
address = {New York, NY, USA},
url = {https://doi.org/10.1145/1553374.1553380},
doi = {10.1145/1553374.1553380},
booktitle = {Proceedings of the 26th Annual International Conference on Machine Learning},
pages = {41–48},
numpages = {8},
location = {Montreal, Quebec, Canada},
series = {ICML '09}
}

@misc{yang2026ttcstesttimecurriculumsynthesis,
      title={TTCS: Test-Time Curriculum Synthesis for Self-Evolving}, 
      author={Chengyi Yang and Zhishang Xiang and Yunbo Tang and Zongpei Teng and Chengsong Huang and Fei Long and Yuhan Liu and Jinsong Su},
      year={2026},
      eprint={2601.22628},
      archivePrefix={arXiv},
      primaryClass={cs.LG},
      url={https://arxiv.org/abs/2601.22628}, 
}

@InProceedings{pmlr-v80-weinshall18a,
  title = 	 {Curriculum Learning by Transfer Learning: Theory and Experiments with Deep Networks},
  author =       {Weinshall, Daphna and Cohen, Gad and Amir, Dan},
  booktitle = 	 {Proceedings of the 35th International Conference on Machine Learning},
  pages = 	 {5238--5246},
  year = 	 {2018},
  editor = 	 {Dy, Jennifer and Krause, Andreas},
  volume = 	 {80},
  series = 	 {Proceedings of Machine Learning Research},
  month = 	 {10--15 Jul},
  publisher =    {PMLR},
  url = 	 {https://proceedings.mlr.press/v80/weinshall18a.html}
}

@inproceedings{xu-etal-2020-curriculum,
    title = "Curriculum Learning for Natural Language Understanding",
    author = "Xu, Benfeng  and
      Zhang, Licheng  and
      Mao, Zhendong  and
      Wang, Quan  and
      Xie, Hongtao  and
      Zhang, Yongdong",
    editor = "Jurafsky, Dan  and
      Chai, Joyce  and
      Schluter, Natalie  and
      Tetreault, Joel",
    booktitle = "Proceedings of the 58th Annual Meeting of the Association for Computational Linguistics",
    month = jul,
    year = "2020",
    address = "Online",
    publisher = "Association for Computational Linguistics",
    url = "https://aclanthology.org/2020.acl-main.542/",
    doi = "10.18653/v1/2020.acl-main.542",
    pages = "6095--6104"
}

@ARTICLE{9392296,
  author={Wang, Xin and Chen, Yudong and Zhu, Wenwu},
  journal={IEEE Transactions on Pattern Analysis and Machine Intelligence}, 
  title={A Survey on Curriculum Learning}, 
  year={2022},
  volume={44},
  number={9},
  pages={4555-4576},
  doi={10.1109/TPAMI.2021.3069908}}

@misc{kim2024strategicdataorderingenhancing,
      title={Strategic Data Ordering: Enhancing Large Language Model Performance through Curriculum Learning}, 
      author={Jisu Kim and Juhwan Lee},
      year={2024},
      eprint={2405.07490},
      archivePrefix={arXiv},
      primaryClass={cs.CL},
      url={https://arxiv.org/abs/2405.07490}, 
}

@inproceedings{jia-etal-2026-makes,
    title = "What Makes a Good Curriculum? Disentangling the Effects of Data Ordering on {LLM} Mathematical Reasoning",
    author = "Jia, Yaning  and
      Zhang, Chunhui  and
      Diao, Xingjian  and
      Yuan, Xiangchi  and
      Ouyang, Zhongyu  and
      Ma, Chiyu  and
      Vosoughi, Soroush",
    editor = "Liakata, Maria  and
      Moreira, Viviane P.  and
      Zhang, Jiajun  and
      Jurgens, David",
    booktitle = "Proceedings of the 64th Annual Meeting of the {A}ssociation for {C}omputational {L}inguistics (Volume 1: Long Papers)",
    month = jul,
    year = "2026",
    address = "San Diego, California, United States",
    publisher = "Association for Computational Linguistics",
    url = "https://aclanthology.org/2026.acl-long.1591/",
    doi = "10.18653/v1/2026.acl-long.1591",
    pages = "34472--34488",
    ISBN = "979-8-89176-390-6"
}

@inproceedings{hu-etal-2026-fine,
    title = "Fine-Grained Data Ordering Improves Fine-Tuning for Large Language Models",
    author = "Hu, Xiaomeng  and
      Tang, Yixuan  and
      Li, Haoze  and
      Chen, Hao  and
      Zhang, Qi  and
      Shen, Zhanming  and
      Zhang, Yiming  and
      Wang, Haobo  and
      Zhao, Junbo",
    editor = "Liakata, Maria  and
      Moreira, Viviane P.  and
      Zhang, Jiajun  and
      Jurgens, David",
    booktitle = "Findings of the {A}ssociation for {C}omputational {L}inguistics: {ACL} 2026",
    month = jul,
    year = "2026",
    address = "San Diego, California, United States",
    publisher = "Association for Computational Linguistics",
    url = "https://aclanthology.org/2026.findings-acl.1021/",
    doi = "10.18653/v1/2026.findings-acl.1021",
    pages = "20406--20418",
    ISBN = "979-8-89176-395-1"
}

@inproceedings{ranaldi-etal-2024-language,
    title = "Does the \textit{Order} Matter? {C}urriculum Learning over Languages",
    author = "Ranaldi, Leonardo  and
      Pucci, Giulia  and
      Freitas, Andr{\`e}",
    editor = "Calzolari, Nicoletta  and
      Kan, Min-Yen  and
      Hoste, Veronique  and
      Lenci, Alessandro  and
      Sakti, Sakriani  and
      Xue, Nianwen",
    booktitle = "Proceedings of the 2024 Joint International Conference on Computational Linguistics, Language Resources and Evaluation (LREC-COLING 2024)",
    month = may,
    year = "2024",
    address = "Torino, Italia",
    publisher = "ELRA and ICCL",
    url = "https://aclanthology.org/2024.lrec-main.464/",
    pages = "5212--5220"
}

@inproceedings{zhang-etal-2026-beyond,
    title = "Beyond Random Sampling: Efficient Language Model Pretraining via Curriculum Learning",
    author = "Zhang, Yang  and
      Mohamed, Amr  and
      Abdine, Hadi  and
      Shang, Guokan  and
      Vazirgiannis, Michalis",
    editor = "Demberg, Vera  and
      Inui, Kentaro  and
      Marquez, Llu{\'i}s",
    booktitle = "Proceedings of the 19th Conference of the {E}uropean Chapter of the {A}ssociation for {C}omputational {L}inguistics (Volume 1: Long Papers)",
    month = mar,
    year = "2026",
    address = "Rabat, Morocco",
    publisher = "Association for Computational Linguistics",
    url = "https://aclanthology.org/2026.eacl-long.271/",
    doi = "10.18653/v1/2026.eacl-long.271",
    pages = "5776--5794",
    ISBN = "979-8-89176-380-7"
}

@inproceedings{ICLR2025_9c77f2ce,
 author = {PENG, XIANGYU and Xia, Congying and Yang, Xinyi and Xiong, Caiming and Wu, Chien-Sheng and Xing, Chen},
 booktitle = {International Conference on Learning Representations},
 editor = {Y. Yue and A. Garg and N. Peng and F. Sha and R. Yu},
 pages = {62484--62519},
 title = {ReGenesis: LLMs can Grow into Reasoning Generalists via Self-Improvement},
 url = {https://proceedings.iclr.cc/paper_files/paper/2025/file/9c77f2ce42151b2c2e26d2cf47f99564-Paper-Conference.pdf},
 volume = {2025},
 year = {2025}
}

@inproceedings{10.1145/3589335.3641257,
author = {Wang, Xin and Zhou, Yuwei and Chen, Hong and Zhu, Wenwu},
title = {Curriculum Learning: Theories, Approaches, Applications, Tools, and Future Directions in the Era of Large Language Models},
year = {2024},
isbn = {9798400701726},
publisher = {Association for Computing Machinery},
address = {New York, NY, USA},
url = {https://doi.org/10.1145/3589335.3641257},
doi = {10.1145/3589335.3641257},
booktitle = {Companion Proceedings of the ACM Web Conference 2024},
pages = {1306–1310},
numpages = {5},
location = {Singapore, Singapore},
series = {WWW '24}
}

@misc{wu2025progressivemasterycustomizedcurriculum,
      title={Progressive Mastery: Customized Curriculum Learning with Guided Prompting for Mathematical Reasoning}, 
      author={Muling Wu and Qi Qian and Wenhao Liu and Xiaohua Wang and Zisu Huang and Di Liang and LI Miao and Shihan Dou and Changze Lv and Zhenghua Wang and Zhibo Xu and Lina Chen and Tianlong Li and Xiaoqing Zheng and Xuanjing Huang},
      year={2025},
      eprint={2506.04065},
      archivePrefix={arXiv},
      primaryClass={cs.CL},
      url={https://arxiv.org/abs/2506.04065}, 
}

@misc{soviany2022curriculumlearningsurvey,
      title={Curriculum Learning: A Survey}, 
      author={Petru Soviany and Radu Tudor Ionescu and Paolo Rota and Nicu Sebe},
      year={2022},
      eprint={2101.10382},
      archivePrefix={arXiv},
      primaryClass={cs.LG},
      url={https://arxiv.org/abs/2101.10382}, 
}

@article{JMLR:v21:20-212,
  author  = {Sanmit Narvekar and Bei Peng and Matteo Leonetti and Jivko Sinapov and Matthew E. Taylor and Peter Stone},
  title   = {Curriculum Learning for Reinforcement Learning Domains: A Framework and Survey},
  journal = {Journal of Machine Learning Research},
  year    = {2020},
  volume  = {21},
  number  = {181},
  pages   = {1--50},
  url     = {http://jmlr.org/papers/v21/20-212.html}
}

@article{Jiang_Meng_Zhao_Shan_Hauptmann_2015, title={Self-Paced Curriculum Learning}, volume={29}, url={https://ojs.aaai.org/index.php/AAAI/article/view/9608}, DOI={10.1609/aaai.v29i1.9608}, abstractNote={ &amp;lt;p&amp;gt; Curriculum learning (CL) or self-paced learning (SPL) represents a recently proposed learning regime inspired by the learning process of humans and animals that gradually proceeds from easy to more complex samples in training. The two methods share a similar conceptual learning paradigm, but differ in specific learning schemes. In CL, the curriculum is predetermined by prior knowledge, and remain fixed thereafter. Therefore, this type of method heavily relies on the quality of prior knowledge while ignoring feedback about the learner. In SPL, the curriculum is dynamically determined to adjust to the learning pace of the leaner. However, SPL is unable to deal with prior knowledge, rendering it prone to overfitting. In this paper, we discover the missing link between CL and SPL, and propose a unified framework named self-paced curriculum leaning (SPCL). SPCL is formulated as a concise optimization problem that takes into account both prior knowledge known before training and the learning progress during training. In comparison to human education, SPCL is analogous to &amp;quot;instructor-student-collaborative&amp;quot; learning mode, as opposed to &amp;quot;instructor-driven&amp;quot; in CL or &amp;quot;student-driven&amp;quot; in SPL. Empirically, we show that the advantage of SPCL on two tasks. &amp;lt;/p&amp;gt; }, number={1}, journal={Proceedings of the AAAI Conference on Artificial Intelligence}, author={Jiang, Lu and Meng, Deyu and Zhao, Qian and Shan, Shiguang and Hauptmann, Alexander}, year={2015}, month={Feb.} }

@ARTICLE{8278851,
  author={Ren, Zhipeng and Dong, Daoyi and Li, Huaxiong and Chen, Chunlin},
  journal={IEEE Transactions on Neural Networks and Learning Systems}, 
  title={Self-Paced Prioritized Curriculum Learning With Coverage Penalty in Deep Reinforcement Learning}, 
  year={2018},
  volume={29},
  number={6},
  pages={2216-2226},
  doi={10.1109/TNNLS.2018.2790981}}

@inproceedings{wang-etal-2026-dsmentor,
    title = "{DSM}entor: Curriculum-Guided Inference with Online Memory for Data-Science {LLM} Agents",
    author = "Wang, He  and
      Li, Alexander Hanbo  and
      Hu, Yiqun  and
      Zhang, Sheng  and
      Kobayashi, Hideo  and
      Zhang, Jiani  and
      Zhu, Henghui  and
      Hang, Chung-Wei  and
      Ng, Patrick",
    editor = "Gupta, Vivek  and
      Ding, Kaize  and
      Kokel, Harsha  and
      Zhao, Yue  and
      Agarwal, Amit  and
      Wang, Yu  and
      Glass, Michael  and
      Zhang, Yu  and
      Srinivas, Kavitha  and
      Chen, Xiusi  and
      Hassanzadeh, Oktie  and
      Zhu, Qi  and
      Chang, Shuaichen  and
      Luo, Yuan",
    booktitle = "Proceedings of the First Workshop on Structured Understanding, Retrieval, and Generation in the {LLM} Era ({SURG}e{LLM} 2026)",
    month = jul,
    year = "2026",
    address = "San Diego, California, United States",
    publisher = "Association for Computational Linguistics",
    url = "https://aclanthology.org/2026.surgellm-1.12/",
    doi = "10.18653/v1/2026.surgellm-1.12",
    pages = "190--208",
    ISBN = "979-8-89176-406-4"
}

@misc{rampp2024doesdefinitiondifficultymatter,
      title={Does the Definition of Difficulty Matter? Scoring Functions and their Role for Curriculum Learning}, 
      author={Simon Rampp and Manuel Milling and Andreas Triantafyllopoulos and Björn W. Schuller},
      year={2024},
      eprint={2411.00973},
      archivePrefix={arXiv},
      primaryClass={cs.LG},
      url={https://arxiv.org/abs/2411.00973}, 
}

@article{WONG2026108438,
title = {Robust heterogeneous network representation learning by multifaceted curriculum training},
journal = {Neural Networks},
volume = {196},
pages = {108438},
year = {2026},
issn = {0893-6080},
doi = {https://doi.org/10.1016/j.neunet.2025.108438},
url = {https://www.sciencedirect.com/science/article/pii/S089360802501319X},
author = {Zhen Hao Wong and Hansi Yang and Quanming Yao and Yaqing Wang}
}

@inproceedings{tao2026dynamic,
  title={Dynamic Curriculum Learning over Difficulty Heterogeneity},
  author={Tao, Chongyang and Xing$^1$, Shihao},
  booktitle={Knowledge Science, Engineering and Management: 19th International Conference, KSEM 2026, Beijing, China, July 17--19, 2026, Proceedings, Part I},
  pages={158},
  year={2026},
  organization={Springer Nature}
}

@misc{xia2024lessselectinginfluentialdata,
      title={LESS: Selecting Influential Data for Targeted Instruction Tuning}, 
      author={Mengzhou Xia and Sadhika Malladi and Suchin Gururangan and Sanjeev Arora and Danqi Chen},
      year={2024},
      eprint={2402.04333},
      archivePrefix={arXiv},
      primaryClass={cs.CL},
      url={https://arxiv.org/abs/2402.04333}, 
}

@article{Hammoudeh_2024,
   title={Training data influence analysis and estimation: a survey},
   volume={113},
   ISSN={1573-0565},
   url={http://dx.doi.org/10.1007/s10994-023-06495-7},
   DOI={10.1007/s10994-023-06495-7},
   number={5},
   journal={Machine Learning},
   publisher={Springer Science and Business Media LLC},
   author={Hammoudeh, Zayd and Lowd, Daniel},
   year={2024},
   month=Mar, pages={2351–2403} }

@InProceedings{pmlr-v151-silva22a,
  title = 	 { Cross-Loss Influence Functions to Explain Deep Network Representations },
  author =       {Silva, Andrew and Chopra, Rohit and Gombolay, Matthew},
  booktitle = 	 {Proceedings of The 25th International Conference on Artificial Intelligence and Statistics},
  pages = 	 {1--17},
  year = 	 {2022},
  editor = 	 {Camps-Valls, Gustau and Ruiz, Francisco J. R. and Valera, Isabel},
  volume = 	 {151},
  series = 	 {Proceedings of Machine Learning Research},
  month = 	 {28--30 Mar},
  publisher =    {PMLR},
  url = 	 {https://proceedings.mlr.press/v151/silva22a.html}
}

@InProceedings{pmlr-v267-wang25bm,
  title = 	 {{NICE} Data Selection for Instruction Tuning in {LLM}s with Non-differentiable Evaluation Metric},
  author =       {Wang, Jingtan and Lin, Xiaoqiang and Qiao, Rui and Koh, Pang Wei and Foo, Chuan-Sheng and Low, Bryan Kian Hsiang},
  booktitle = 	 {Proceedings of the 42nd International Conference on Machine Learning},
  pages = 	 {63662--63689},
  year = 	 {2025},
  editor = 	 {Singh, Aarti and Fazel, Maryam and Hsu, Daniel and Lacoste-Julien, Simon and Berkenkamp, Felix and Maharaj, Tegan and Wagstaff, Kiri and Zhu, Jerry},
  volume = 	 {267},
  series = 	 {Proceedings of Machine Learning Research},
  month = 	 {13--19 Jul},
  publisher =    {PMLR},
  url = 	 {https://proceedings.mlr.press/v267/wang25bm.html}
}

@inproceedings{NEURIPS2025_cea04322,
 author = {Zhang, Dylan and Dai, Qirun and Peng, Hao},
 booktitle = {Advances in Neural Information Processing Systems},
 editor = {D. Belgrave and C. Zhang and H. Lin and R. Pascanu and P. Koniusz and M. Ghassemi and N. Chen},
 pages = {141172--141208},
 publisher = {Curran Associates, Inc.},
 title = {The Best Instruction-Tuning Data are Those That Fit},
 url = {https://proceedings.neurips.cc/paper_files/paper/2025/file/cea04322465ad2f261f08e5b47ba9e7a-Paper-Conference.pdf},
 volume = {38},
 year = {2025}
}

@inproceedings{NEURIPS2025_a59ff5f7,
 author = {Fu, Yanjun and Hamman, Faisal and Dutta, Sanghamitra},
 booktitle = {Advances in Neural Information Processing Systems},
 editor = {D. Belgrave and C. Zhang and H. Lin and R. Pascanu and P. Koniusz and M. Ghassemi and N. Chen},
 pages = {113932--113958},
 publisher = {Curran Associates, Inc.},
 title = {T-SHIRT: Token-Selective Hierarchical Data Selection for Instruction Tuning},
 url = {https://proceedings.neurips.cc/paper_files/paper/2025/file/a59ff5f7384176b5d14a5ded77c4aa4f-Paper-Conference.pdf},
 volume = {38},
 year = {2025}
}

@misc{hui2024qwen25codertechnicalreport,
      title={Qwen2.5-Coder Technical Report}, 
      author={Binyuan Hui and Jian Yang and Zeyu Cui and Jiaxi Yang and Dayiheng Liu and Lei Zhang and Tianyu Liu and Jiajun Zhang and Bowen Yu and Keming Lu and Kai Dang and Yang Fan and Yichang Zhang and An Yang and Rui Men and Fei Huang and Bo Zheng and Yibo Miao and Shanghaoran Quan and Yunlong Feng and Xingzhang Ren and Xuancheng Ren and Jingren Zhou and Junyang Lin},
      year={2024},
      eprint={2409.12186},
      archivePrefix={arXiv},
      primaryClass={cs.CL},
      url={https://arxiv.org/abs/2409.12186}, 
}

@misc{grattafiori2024llama3herdmodels,
      title={The Llama 3 Herd of Models}, 
      author={Aaron Grattafiori et.al.},
      year={2024},
      eprint={2407.21783},
      archivePrefix={arXiv},
      primaryClass={cs.AI},
      url={https://arxiv.org/abs/2407.21783}, 
}

@inproceedings{xu-etal-2025-kodcode,
    title = "{K}od{C}ode: A Diverse, Challenging, and Verifiable Synthetic Dataset for Coding",
    author = "Xu, Zhangchen  and
      Liu, Yang  and
      Yin, Yueqin  and
      Zhou, Mingyuan  and
      Poovendran, Radha",
    editor = "Che, Wanxiang  and
      Nabende, Joyce  and
      Shutova, Ekaterina  and
      Pilehvar, Mohammad Taher",
    booktitle = "Findings of the Association for Computational Linguistics: ACL 2025",
    month = jul,
    year = "2025",
    address = "Vienna, Austria",
    publisher = "Association for Computational Linguistics",
    url = "https://aclanthology.org/2025.findings-acl.365/",
    doi = "10.18653/v1/2025.findings-acl.365",
    pages = "6980--7008",
    ISBN = "979-8-89176-256-5"
}

@inproceedings{ICLR2025_c239bac7,
 author = {Ye, Tian and Xu, Zicheng and Li, Yuanzhi and Allen-Zhu, Zeyuan},
 booktitle = {International Conference on Learning Representations},
 editor = {Y. Yue and A. Garg and N. Peng and F. Sha and R. Yu},
 pages = {78136--78147},
 title = {Physics of Language Models: Part 2.2, How to Learn From Mistakes on Grade-School Math Problems},
 url = {https://proceedings.iclr.cc/paper_files/paper/2025/file/c239bac713017b0b2257b7622bf8aab3-Paper-Conference.pdf},
 volume = {2025},
 year = {2025}
}






\end{document}